# Interpretable machine learning-accelerated seed treatment by nanomaterials for environmental stress alleviation

Hengjie Yu[a,b], Dan Luo[c], Sam F. Y. Li[d], Maozhen Qu[a,b], Da Liu[a,b], Yingchao He[a,b], and Fang Cheng[a,b*].

[a]College of Biosystems Engineering and Food Science, Zhejiang University, Hangzhou 310058, China

[b]Key Laboratory of Intelligent Equipment and Robotics for Agriculture of Zhejiang Province, Hangzhou 310058, China

[c]Department of Biological and Environmental Engineering, Cornell University, Ithaca, New York 14853, USA

[d]Department of Chemistry, National University of Singapore, 3 Science Drive 3, Singapore 117543, Singapore

***Email:** fcheng@zju.edu.cn

## Abstract

Crops are constantly challenged by different environmental conditions. Seed treatment by nanomaterials is a cost-effective and environmentally-friendly solution for environmental stress mitigation in crop plants. Here, 56 seed nanopriming treatments are used to alleviate environmental stresses in maize. Seven selected nanopriming treatments significantly increase the stress resistance index (SRI) by 13.9% and 12.6% under salinity stress and combined heat-drought stress, respectively. Metabolomics data reveals that ZnO nanopriming treatment, with the highest SRI value, mainly regulates the pathways of amino acid metabolism, secondary metabolite synthesis, carbohydrate metabolism, and translation. Understanding the mechanism of seed nanopriming is still difficult due to the variety of nanomaterials and the complexity of interactions between nanomaterials and plants. Using the nanopriming data, we present an interpretable structure-activity relationship (ISAR) approach based on interpretable machine learning for predicting and understanding its stress mitigation effects. The post hoc and model-based interpretation approaches of machine learning are combined to provide complementary benefits and give researchers or policymakers more illuminating or trustworthy results. The concentration, size, and zeta potential of nanoparticles are identified as dominant factors for correlating root dry weight under salinity stress, and their effects and interactions are explained. Additionally, a web-based interactive tool is developed for offering prediction-level interpretation and gathering more details about specific nanopriming treatments. This work offers a promising framework for accelerating the agricultural applications of nanomaterials and may profoundly contribute to nanosafety assessment.

**Keywords:** environmental stress, metal oxide nanomaterials, seed nanopriming, machine learning, model interpretation

## 1. Introduction

Crops are constantly challenged by different environmental conditions, such as drought, salinity, and extreme temperatures (Ahuja et al., 2010; Ioannou et al., 2020). Various field applications of nanoparticles were tested to mitigate environmental stresses (Kah et al., 2019; Zhao et al., 2020), such as drought stress in maize (Cu nanoparticles) (Van Nguyen et al., 2022), high-temperature stress in grain sorghum (Se nanoparticles) (Djanaguiraman et al., 2018), and salinity stress in broad bean ($TiO_2$ nanoparticles) (Abdel Latef et al., 2018). Compared with traditional analogs, it was estimated that the median efficiency of nanoagrochemicals increased by ~20-30% (Kah et al., 2018). Seed priming by nanomaterials is a novel approach that is cost-effective and environmentally-friendly due to the lower amount of nanomaterials and minimized environmental exposure compared to field applications (De La Torre-Roche et al., 2020; Hofmann et al., 2020). Seed nanopriming could stimulate the growth of crop plants under environmental stresses by increasing photosynthetic pigments levels (Abdel Latef et al., 2017), enhancing antioxidant enzyme activities (Shah et al., 2021), reducing reactive oxygen species (ROS) production (Rai-Kalal et al., 2021), and regulating the $H_2O_2$ signaling network (Rai-Kalal et al., 2021). Although most current nanopriming studies, to the best of our knowledge, focus only on single environmental stress (Hussain et al., 2019; Khan et al., 2020; Rizwan et al., 2019), we hypothesized that seed nanopriming might have great potential to alleviate multiple environmental stresses due to several underlying common mechanisms.

Quantitative/qualitative structure-activity (QSAR) relationships are well-established computational approaches for chemical design and early hazard warnings (Muratov et al., 2020; Nongonierma et al., 2018). Machine learning, suitable for addressing ill-defined problems where nonlinearities and interactions exist, is employed to establish relationships between descriptors and a target. The interpretation of good-performance models, however, is challenging for extracting information and building trust. Interpretable machine learning receives increasing attention due to its ability to provide insight into what a model has learned, whose interpretation techniques could be organized into model-based and post hoc categories (Murdoch et al., 2019). Model-based interpretation usually is generated by simple and interpretable models, which may result in lower prediction accuracy. It is worth noting that few self-interpretable models could achieve acceptable performance but with unintuitive model interpretation, such as a large tree depth in the decision tree model and numerous decision rules in the RuleFit algorithm. Post hoc interpretation methods are conducted to describe the learned relationships by trained models instead of changing the models, which may be based on sophisticated models with good performance. Many post hoc interpretation methods, however, are relatively new, with much of questions to be answered. Moreover, it is challenging to determine which method could fully capture a model's behavior. The integration of different interpretation methods is therefore preferred, including the integration of different methods for one interpretation task and the integration of model-based and post hoc interpretation.

In this study, a variety of metalloid and metal oxide nanoparticles ($SiO_2$, $CeO_2$, $CuO$, $Fe_3O_4$, $ZnO$,

α-Fe$_2$O$_3$, and γ-Fe$_2$O$_3$ of different sizes) were used as priming agents with different concentrations for alleviating environmental stress in maize; maize is chosen due to the fact that it is one of the most important staple crops. Salinity, drought, and heat stresses are major constraints to maize yield production (Yuan et al., 2019; Zhao et al., 2017). Besides, oxide nanoparticles are among the widest-used manufactured nanomaterials (Chavali and Nikolova, 2019; Tsugita et al., 2017), and they can be acquired at a relatively little cost which is very important for agricultural use. A total of fourteen oxide nanomaterials were tested with different concentrations. Eleven biological endpoints were collected for investigating the effects of seed nanopriming on germination under salinity stress. Then, the treatments with good performance were selected to further investigate their effects under combined heat-drought stress. Metabolomics analysis was used to study small-molecule metabolites and provide insight into the mechanisms of seed nanopriming for stress mitigation. Furthermore, using the seed nanopriming data under salinity stress, we presented an interpretable structure-activity relationship (ISAR) approach based on the integration of machine learning interpretation methods to identify important factors and show how they affect the stress mitigation effects. Light gradient boosting machine (LightGBM), RuleFit, and decision tree models were established as complex and self-interpretable models for providing post hoc and model-based interpretation. Moreover, we deployed the established models and their local interpretation in an online interactive website where interested readers could learn more about the specific nanopriming treatment (or a customized treatment) based on the prediction-level interpretation.

**2. Materials and Methods**

**2.1 Materials and chemicals**

ZnO (30±10 and 50±10 nm), CeO$_2$ (20-50 nm), TiO$_2$ (20 and 40 nm), CuO (40 nm), α-Fe$_2$O$_3$ (30 nm), Fe$_3$O$_4$ (20 nm), and SiO$_2$ (20±5 and 50±5 nm) nanoparticles were purchased from Shanghai Macklin Biochemical Co., Ltd. CuO (50-100 nm), CeO$_2$ (<100 nm) and γ- Fe$_2$O$_3$ (≤50 nm) nanoparticles were obtained from Shanghai Aladdin Reagent Co., Ltd. Fe$_3$O$_4$ (50 nm) nanoparticles were purchased from Shanghai Yien Chemical Technology Co., Ltd. NaCl was obtained from Sinopharm Chemical Reagent Co., Ltd. Ammonium acetate (NH$_4$Ac), ammonium hydroxide ($NH_4OH$), and acetonitrile was purchased from Merck & Co., Inc. Maize seeds (cultivar Jixiang Yihao) were obtained from Gansu Dunhuang Seed Group Co., Ltd. The average germination rate was >90% in a trial study.

**2.2. Characterization of nanoparticles**

TEM characterization was used to obtain nanoparticle size and morphology on a JEOL JEM-1400Flash. The hydrodynamic diameter (Z-average), polydispersity index (PdI), and zeta potential were measured using Malvern Zetasizer Nano ZS90 based on dynamic light scattering (DLS) and laser doppler velocimetry (LDV) techniques. The nanoparticles were dispersed using ultrasonic vibration (240 W, 40 kHz) for 30 min. Samples were prepared for TEM characterization by dipping copper mesh in suspension and drying it at room temperature. After standing at room temperature for 48 h, the supernatants of nanosuspensions were used for DLS and LDV measurement. The TEM images were shown in Fig. S1, and the TEM size was obtained after counting the sizes of more than 100 nanoparticles.

**2.3. Seed nanopriming treatment**

The maize seeds with a width of over 9.5 mm and a thickness of over 5.5 mm were first selected by a round hole sieve and a slotted hole sieve. Then the seeds with any morphological damage or discoloration were removed. The selected seeds were immersed in 70% ethyl alcohol for 2 min, followed by 5% sodium hypochlorite for 10 min to ensure surface sterility, and then washed thoroughly with deionized (DI) water. Sterilized seeds were dried at room temperature to their original weight. The 500 mg/L nanoparticle suspensions were prepared and then dispersed using ultrasonic vibration (240 W, 40 kHz) for 30 min. Then the nanosuspensions were diluted to different concentrations (25, 50, 100, 200 mg/L) for seed nanopriming. The seeds were primed in nanoparticle suspensions in centrifuge tubes for 12 h in the dark, and the tubes were placed horizontally in a shaker (150 r/min, 26±1°C). The proportion of seed weight to suspension volume was 1:5 g/mL. The $H_2O$ priming treatment was used as a control. The primed seeds were washed with DI water three times for 3 min each time, followed by drying at room temperature to their original weight. The dried seeds were stored in a refrigerator at 4°C.

**2.4. Seed germination assay under salinity and combined heat-drought stresses**

Eight seeds were planted at 2 cm depth in a square pot (10 cm×10 cm×7 cm) filled with about 265 g of washed and air-dried vermiculite. The 100 mL of 1/4 Hogland nutrient solution was added after planting. Then 40 mL of 1/4 Hogland nutrient solution was daily added for the next 3 days and 1/2 Hogland nutrient solution afterward (Hoagland et al., 1950). The used nutrient solutions used on the 1st, 3rd, 5th, and 7th day contained 150 mM NaCl for salinity stress. Three replicates were applied for each treatment. The square pots were placed in a climate chamber with a 14-10h photoperiod (26/22°C day/night temperature, 60% relative humidity, 45,000 lx light intensity), and their positions were changed every day.

The treatments with better performance in mitigating effects under salinity stress were selected to investigate their effects under combined heat-drought stress. The sowing and nutrient solution use were consistent with salinity stress but without NaCl treatment. After four-day cultivation, the parameters of the climate chamber were changed to 40/32°C day/night temperature and 50% relative humidity. Although the amount of daily watering was not changed, the plants faced drought stress due to increased evaporation of water. After another three days, the plants were harvested and the values of eleven biological endpoints were collected.

**2.5. Biological endpoint collection and statistical analysis**

After seven-day cultivation, eleven biological endpoints, including chlorophyll content, leaf area, shoot length, root/stem/leaf fresh/dry weight, root:shoot ratio, and SRI, were obtained to reflect stress alleviation in the germination of maize seeds after seed nanopriming. The chlorophyll content in leaves was measured by the SPAD-502 chlorophyll meter about 2 h before harvest. The seedling roots were washed with water to remove the residual culture medium, and the seedlings were kept in ziplock bags and placed in a 4°C refrigerator until further measurements in a few hours. The shoot length was measured using a rule. The leaf area was calculated by multiplying the product of leaf length and width by 0.75. The calculation of root:shoot ratio and SRI is listed in Supplementary methods S1 and S2.

Values of three independent experiments were expressed as means ± standard deviation (SD). The statistical significance among treatments was determined by analysis of variance (ANOVA) and the least significant difference (LSD) test at $p<0.05$ level. Student's t-test was used to compare the values of the two groups, and ns, \*, \*\*, and \*\*\* represented the significance level of non-significant, 0.1, 0.05, and 0.01 respectively. The statistical analysis was conducted based on the R package (agricolae).

**2.6. Metabolite extraction and metabonomics analysis**

The treatment with the best stress mitigation effect, ZnO nanopriming treatment (30nm, 200mg/L), was selected to collect samples for metabonomics analysis under salinity and combined heat-drought stresses respectively following the nanopriming and seedling cultivation methods described above. The maize leaves (first leaves for salinity stress and second leaves for combined heat-drought stress) were collected from seven-day-old plants and frozen in liquid nitrogen immediately for 15 min, then kept at -80°C until use.

The maize tissues (80 mg leaves) retrieved from -80 °C storage were ground into fine powder in liquid nitrogen. The powder was vortexed in a 1 mL solution of methanol/acetonitrile/water (2:2:1, v/v/v) and centrifuged for 20 min (14000 g and 4 °C). The supernatant was then dried in a vacuum centrifuge. A 100 μL solvent of acetonitrile/water (1:1, v/v) was added to re-dissolve the sample and then centrifuged for 15 min (14000 g at 4 °C). The supernatant was then injected for LC-MS analysis.

Analyses were performed using UHPLC (1290 Infinity LC, Agilent Technologies) coupled to a quadrupole time of flight (AB Sciex TripleTOF 6600) in Shanghai Applied Protein Technology Co., Ltd. A random sequence was used for the sample analysis to avoid the influence caused by instrument fluctuation. Quality control (QC) samples were inserted into the sample queue to evaluate the reliability of the data. More details regarding the analytical method, instrument parameters, and data acquisition are provided in Supplementary method S3.

The metabolomics analysis was performed in six replicate samples. The data processing and statistical analysis followed a procedure described in a previous study with some adjustments (Yu et al., 2022). The cenWave m/z was set to 10 ppm, and compound identification of metabolites was performed by comparing of accuracy m/z value (<10 ppm). PLS-DA and OPLS-DA were employed to analyze the normalized data based on the R package (ropls). The VIP value of each variable in the OPLS-DA model was calculated to estimate its contribution to the classification. Student's t-test was employed to determine the significance of differences between two independent samples. Metabolites with a VIP value >1 and p values less than 0.05 were considered statistically significant.

**2.7. Model establishment and model interpretation**

Nine features were collected from experiments, including nanoparticle composition, TEM size, TEM size SD, morphology, concentration, hydrodynamic diameter, PdI, zeta potential, and BET surface area. An overview of the used dataset and the prediction target (root dry weight) is shown in Table S1 and Fig. S2, and the dataset is given in Supplementary dataset S1. Root dry weight was divided into two categories in equal quantities (0 for low level and 1 for high level). Two features (TEM size SD and PdI) were removed due to the high correlation (Fig. S3).

The LightGBM model was employed as a complex model since it is an effective and highly scalable algorithm and can handle categorical features directly. Nonetheless, label encoding is used to encode categorical features for the need of some interpretation methods. Python and two machine learning packages (LightGBM and scikit-learn) were used for classification modeling. The version of the main software and packages used are given in Table S2. The workflow for the establishment of the LightGBM models is shown in Fig. S4. The dataset was split into a training set (75%) and a test set (25%) using the stratified shuffle split method. Grid search and 5-fold CV were used to determine model hyperparameters based on the training set (Table S3). The datasets were randomly split into the training set and test set ten times to evaluate model stability. The average AUROC, F1 score (weighted), and accuracy on the 5-fold CV on the training set and the AUROC, F1 score (weighted), and accuracy on the training set and test set were used to determine the model performance.

Model interpretation workflow, post hoc interpretation, and model-based interpretation were conducted following procedures described in our previous studies with some adjustments (Yu et al., 2021; Yu et al., 2022; Yu et al., 2022). Briefly, Different LightGBM models were established based on ten dataset splits, and LightGBM importance, permutation importance, and SHAP importance were employed to identify important features for ten models. Their average feature importance rank was used to identify the most relevant features for correlating the root dry weight. A model with feature importance consistent with average importance and better model performance was selected for further post hoc interpretation. Partial dependence plot (PDP), individual conditional expectation (ICE), and SHapley Additive exPlanations (SHAP) were employed for the visualization of the feature effects, and the measurement and visualization of feature interactions were achieved by SHAP interaction. Then, the important features identified by post hoc interpretation of the LightGBM model were used for the establishment of self-interpretable models, including decision tree and RuleFit models, which could provide model-based interpretation. Moreover, the post hoc interpretation was conducted to interpret the decision tree model intuitively. The details of line fitting of SHAP main effects and SHAP interactions is listed in Supplementary method S4.

**2.8. Interactive website construction for local interpretation**

An interactive website was constructed using the "streamlit" package based on python. The established LightGBM and decision tree models were deployed on this website, and prediction-level interpretation was obtained in real time after selecting an instance from used datasets or customizing a sample. Moreover, another local interpretation method, LIME, was also provided using the "lime" package.

**3. Results**

**3.1 Statistical analysis of nanopriming treatments on environmental stresses.**

Under salinity stress, the effects of seed nanopriming in germination were concentration-, size-, and composition-dependent. The stress resistance index (SRI) was developed to quantitatively assess the stress tolerance of nanopriming treatments by combining ten physiological parameters of seedlings (Fig. 1A). The average SRI values of seed nanopriming by fourteen types of nanoparticles were higher than that in the control (SRI = 10.0). Among them, the treatments with ZnO (30 nm), ZnO (50 nm), $CeO_2$ (20-50 nm), and $SiO_2$ (50 nm) nanoparticles had the highest average SRI values. For specific nanopriming treatment, the ZnO nanopriming (30 nm at 200 mg/L) had the best result. The readers can get more information about each nanopriming treatment in Fig. 1A.

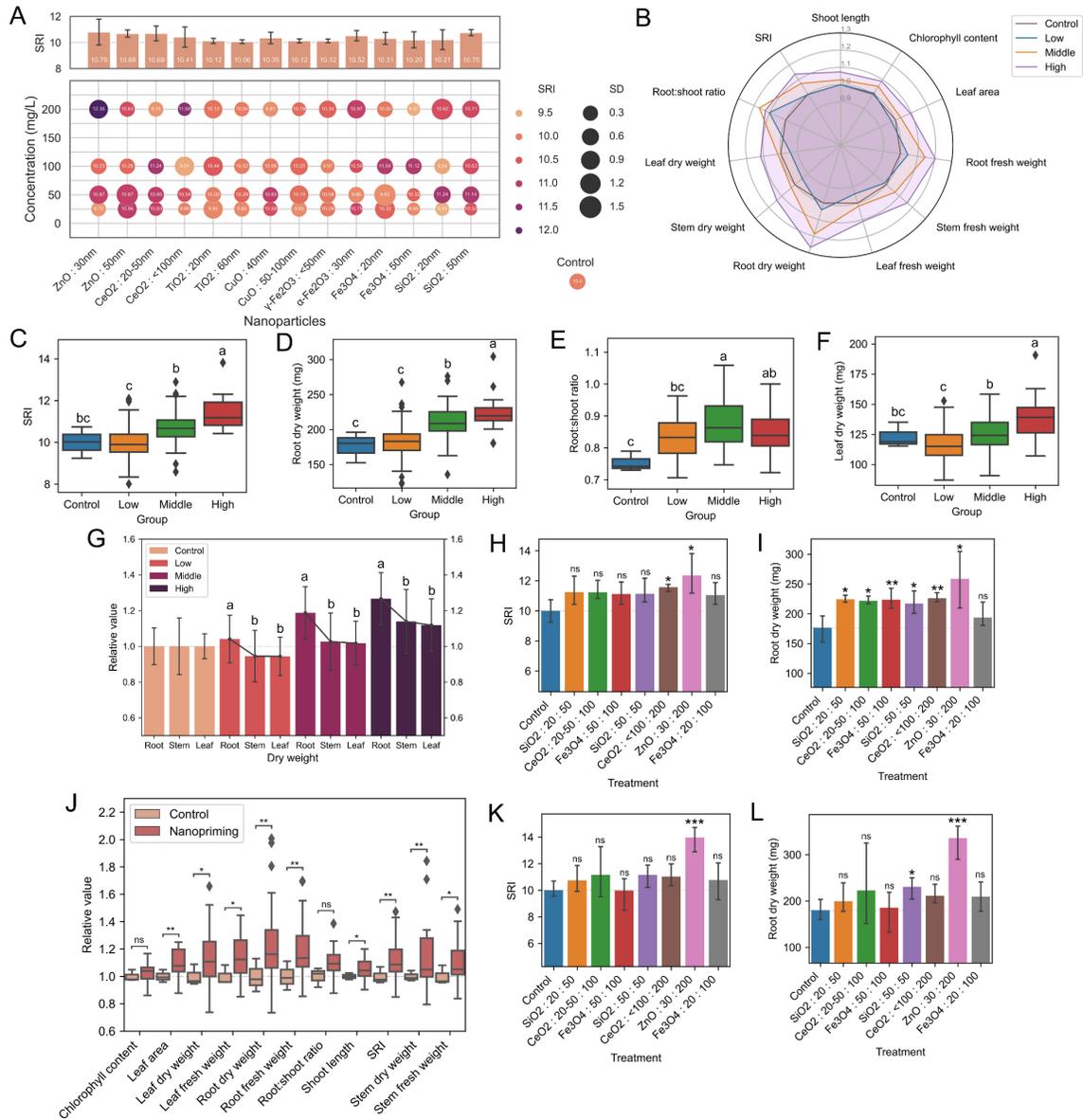

**Fig. 1.** Statistical analysis of nanopriming treatments on salinity and combined heat-drought stresses. A, SRI of 56 seed nanopriming treatments under salinity stress. The error bar and point size represent the SD. The average SRI values for all nanopriming treatments are represented by the color of points. B, The relative values of eleven biological endpoints in three nanopriming groups and the control under salinity stress (divided by the values in the control). C-F, The differences between three nanopriming groups and the control on biological endpoints under salinity stress. G, the comparison among dry weight relative values in the root, stem, and leaf in three nanopriming

groups and the control under salinity stress. H,I, The comparison of biological endpoints among seven selected treatments in the High group under salinity stress. J, The comparison of eleven biological endpoints between selected nanopriming treatments and the control under combined heat-drought stresses. K,L, The comparison of biological endpoints among selected treatments under combined heat-drought stresses.

The 56 nanopriming treatments were divided into three groups based on the SRI values under salinity stress (Table S4), including the High group (SRI≥11), the Middle group (10.5≤SRI<11), and the Low group (SRI<10.5) corresponding to 7, 16, and 33 treatments, respectively. The seven nanopriming treatments in the High group increased the SRI values by 13.9% on average compared with the control. The average values of all endpoints in the High and Middle groups were higher than those in the control (Fig. 1B). There were significant differences ($p<0.05$) in all endpoints between the treatments in the High group and the control; in the Middle group, there were significant differences in five endpoints (including chlorophyll content, root fresh weight, root dry weight, root:shoot ratio, and shoot length) compared with the control (Fig. 1C-F and Fig. S5).

Under salinity stress, root dry weight was the largest improvement of endpoints in the High and Middle groups after seed nanopriming with an increase of 26.7% and 18.7% on average, respectively (Fig. 1B). Root:shoot ratio was the only endpoint that the average value in the Middle group was higher than that in the High group but without significant difference due to the much lower improvement in shoot dry weight in the Middle group. Furthermore, the increase in the root dry weight was significantly higher than that in the stem and leaf dry weight in all nanopriming groups (Fig. 1G).

All the average values of eleven biological endpoints in selected nanopriming treatments were higher than those in the control under salinity stress (Fig. 1H,I and Fig. S6). The treatment of $CeO_2$ (<100 nm and 200 mg/L) was significantly higher than the control in eight biological endpoints. Although the treatment of ZnO (30 nm at 200 mg/L) had the highest SRI value and its average values were higher than the treatment of $CeO_2$ (<100 nm and 200 mg/L) in ten biological endpoints, it was only significantly higher than the control in five biological endpoints because of the large SD in other biological endpoints data.

We selected seven nanopriming treatments in the High group under salinity stress to study their effects under the combined heat-drought stress and to explore the potential of seed nanopriming for the alleviation of a variety of environmental stresses. All median values of biological endpoints of selected nanopriming treatments were higher than the control (Fig. 1J). Furthermore, there were significant differences in nine biological endpoints between the selected treatments and the control except for chlorophyll content and root:shoot ratio. For example, the selected treatments increased the SRI by 12.6% on average. The treatment with the highest SRI under salinity stress, ZnO (30 nm at 200 mg/L) nanopriming, also had the best result under combined heat-drought stress (Fig. 1K). This treatment significantly increased ten biological endpoints, such as SRI (39.8%) and root dry weight (86.4%) (Fig. 1K,L and Fig. S7). Therefore, the results indicated that seed nanopriming could improve seedling growth under both salinity and combined heat-drought stresses, and the root biomass had the highest increase.

**3.2. Metabolomics analysis of ZnO nanopriming treatments on environmental stresses**

Metabolomics is the global analysis of all or wide arrays of metabolites that are regarded as ultimate responses of biological systems to stress(Fiehn, 2002; Shulaev, 2006). An ultra-high performance liquid chromatography/quadrupole time-of-flight mass spectrometry (UHPLC/Q-TOF-MS)-based metabolomics was used to determine the metabolic responses of the best nanopriming treatment, ZnO nanopriming (30 nm at 200 mg/L), under salinity stress and combined heat-drought stress in both positive and negative ion modes. A total of 1204 metabolites were identified, such as lipids and lipid-like molecules (24.92%), phenylpropanoids and polyketides (16.20%), organoheterocyclic compounds (11.46%), benzenoids (9.22%), organic acids and derivatives (8.47%), and organic oxygen compounds (6.40%) (Fig. S8). The ZnO nanopriming treatment under salinity was termed SN, while the control was termed SC; the nanopriming treatment under combined heat-drought stress was termed HdN, while the control was termed HdC. A supervised partial least-squares discriminant analysis (PLS-DA) was performed, and score plots indicated that nanopriming treatments (SN and HdN) were clearly separated from their control groups (Fig. S9). The 7-fold cross-validation (CV) results indicated that the models were stable and reliable (Table S5).

Based on the variable importance in the projection (VIP) value of each variable in the orthogonal partial least-squares discriminant analysis (OPLS-DA) model (VIP > 1) and the p value of Student's t-test ($p < 0.05$), 46 differential metabolites were selected under salinity stress (Fig. 2A and Supplementary dataset S2), of which 35 metabolites were up-regulated and 11 metabolites were down-regulated (Fig. S10). The important metabolic pathways were further analyzed using KEGG (Kyoto Encyclopedia of Genes and Genomes) pathway maps. In salinity stress, significant differences between SN and SC were observed in tryptophan metabolism, phenylalanine, tyrosine and tryptophan biosynthesis, biosynthesis of various secondary metabolites - part 2, and amino sugar and nucleotide sugar metabolism (Fig. 2B and Fig. S11). All the metabolites involved in the differentially enriched pathways were up-regulated, such as anthranilic acid and tryptophan. Under combined heat-drought stress, of 61 differential metabolites (Fig. 2C and Supplementary dataset S3), 44 metabolites were up-regulated while 17 metabolites were down-regulated (Fig. S12). Other significant differences between HdN and HdC (Fig. 2D and Fig. S13) were also evident in the pathways of aminoacyl-tRNA biosynthesis, alanine, aspartate and glutamate metabolism, pantothenate and CoA biosynthesis, beta-Alanine metabolism, biosynthesis of amino acids, ABC transporters, cyanoamino acid metabolism, and amino sugar and nucleotide sugar metabolism. Although over two-thirds of the differential metabolites were up-regulated (such as benzenoids and organic oxygen compounds), most of the metabolites involved in differential metabolic pathways were down-regulated, such as asparagine, valine, histidine, and pantothenic acid.

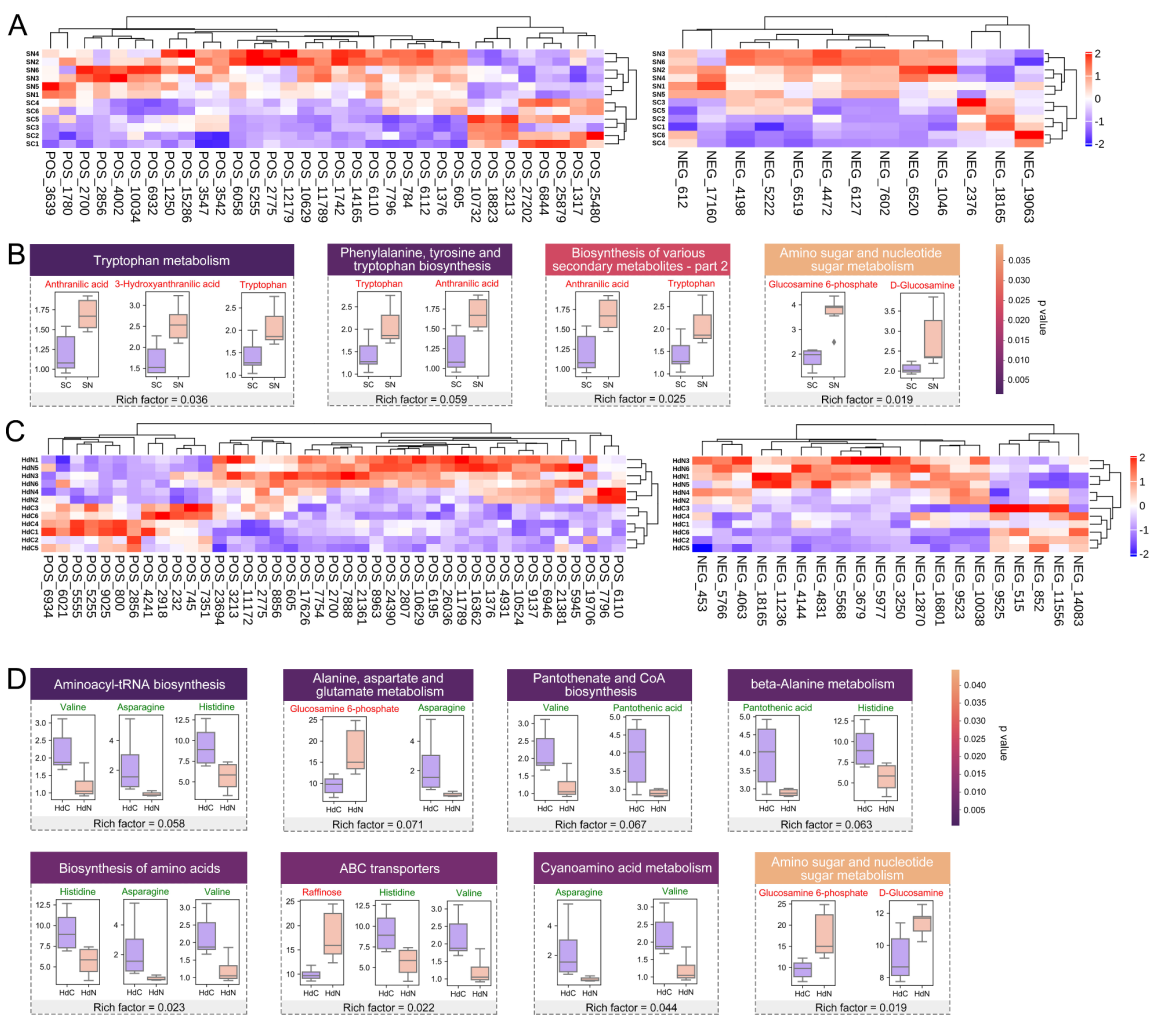

**Fig. 2.** Metabolomics analysis of maize leaves under environmental stresses after seed nanopriming using ZnO nanoparticles (30 nm, 200 mg/L). A, Clustered heatmap of the metabolites in maize leaves under salinity stress. B, Changed metabolites and metabolic pathways in maize leaves under salinity stress. C, Clustered heatmap of the metabolites in maize leaves under combined heat-drought stress. D, Changed metabolites and metabolic pathways in maize leaves under combined heat-drought stress.

**3.3 Interpretation workflow, model establishment, and feature importance**

In order to better understand the mechanism of seed nanopriming, we established ISAR based on the model interpretation of machine learning. Root dry weight was selected for classification as it was the biological endpoint with the greatest improvement and played an important role in the uptake of water and nutrient. The workflow of model interpretation is as follows (Fig. 3A): first post hoc interpretation, next model-based interpretation, and then post hoc interpretation (PMP). LightGBM model was first established as a complex model. The post hoc interpretation of the LightGBM model was conducted to identify important features and show how they affect model output, such as permutation importance, PDP, ICE, and SHAP. Then self-interpretable models (decision tree and RuleFit) were established after the simplification of the problem based on the extracted information from the post hoc interpretation. The model-based interpretation, as we mentioned before, maybe unintuitive due to numerous leaf nodes or decision rules. Therefore, the post hoc interpretation was then conducted for a more intuitive interpretation.

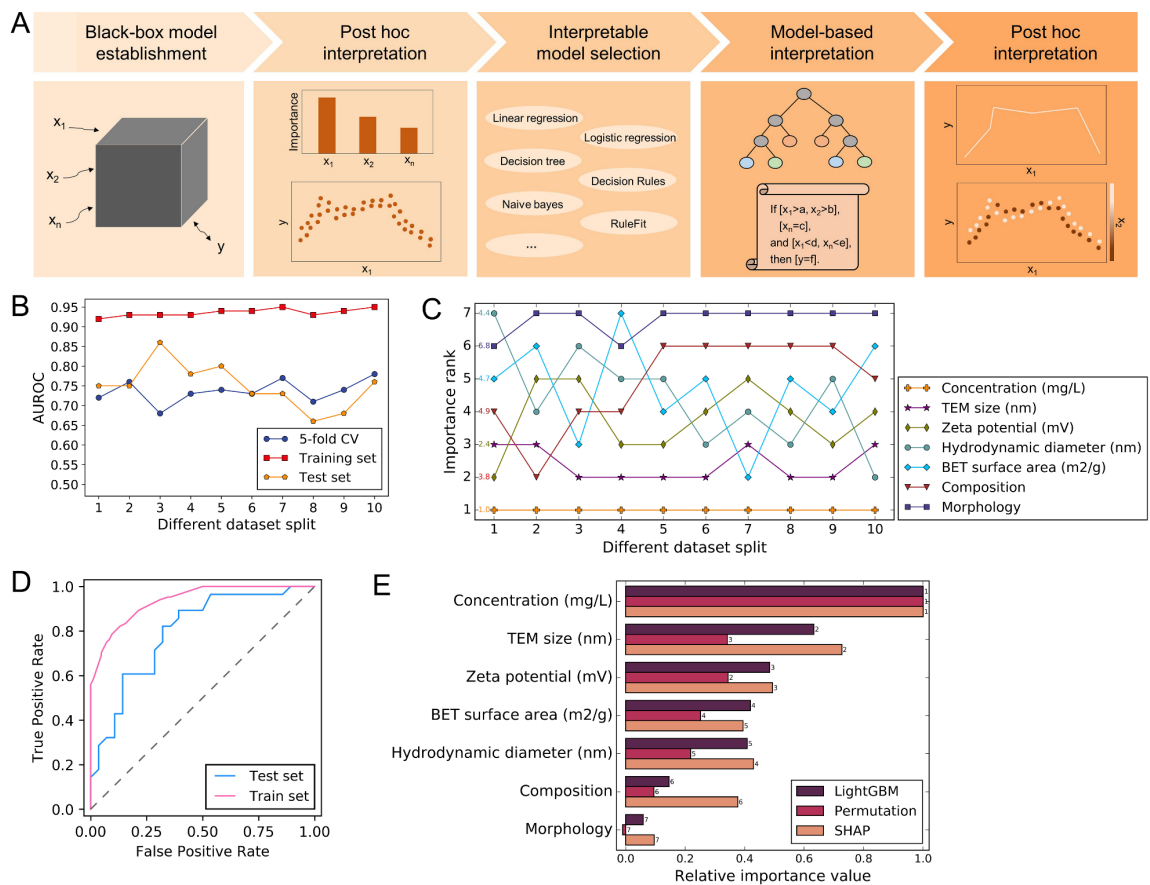

**Fig. 3.** Model establishment, feature importance, and interpretation workflow. A, The workflow of PMP interpretation (the figures are illustrative only, rather than based on data). B, The AUROC of established models on ten dataset splits. C, Average feature importance rank of three methods

(LightGBM, Permutation, and SHAP) on ten dataset splits. The number to the left of the first point is the average feature importance rank. D, ROC curve of the established models trained on the fifth dataset split. E, Relative feature importance of three methods of the established models trained on the fifth dataset split (divided by the maximum value of importance), and the absolute values are provided in Fig. S14. The number to the right of the bar is the importance order of a feature in each method.

The average AUROC, accuracy, and F1 score for the test set on ten different dataset splits were 0.75, 0.72, and 0.73, respectively (Fig. 3B and Fig. S15). Concentration was identified as the most important feature in all dataset splits (Fig. 3C). Size (from TEM) and surface charge (from zeta potential) were always ranked in the top five features, which were also identified as important. The model trained on the fifth dataset split was used for further model interpretation as it had a feature importance ranking similar to the average ranking and had a good performance (Fig. 3D,E).

**3.4 Post hoc interpretation of the LightGBM model**

Feature effects of PDP, ICE, and SHAP approximatively followed the same trends (Fig. 4A-F, Figs. S16 and S17). The concentration of 50 mg/L and 200 mg/L had higher root dry weight. Larger size had higher root dry weight, while it tended to be saturated when TEM size was more than ~20 nm. The effect of zeta potential generally followed a decreasing trend. Two-dimensional PDP showed the combined effects of two features (Fig. 4G-I), similar to the individual effects as the main effects played a major role, which may be used as clues for the custom classification tree. SHAP interaction values were used to measure interaction strength (Fig. 4J). Concentration had the highest interaction value with other features, which may be the reason why the ICE curves were inconsistent (Fig. 4A). Concentration and zeta potential had the strongest interaction, where vertical dispersion existed (Fig. 4K). The interactions between concentration and zeta potential increased the root dry weight when the concentration was 100 mg/L, while it decreased root dry weight when the concentration was 50 mg/L. Besides, the interactions between TEM size and concentration decreased root dry weight in the instances with the concentration of 100 mg/L and 50 mg/L (Fig. 4L). The local interpretation for the ZnO nanopriming treatment with the highest SRI is shown in

Fig. 4M. Six of the seven features played a positive role. For example, the concentration of 200 mg/L greatly pushed the prediction higher than the base value.

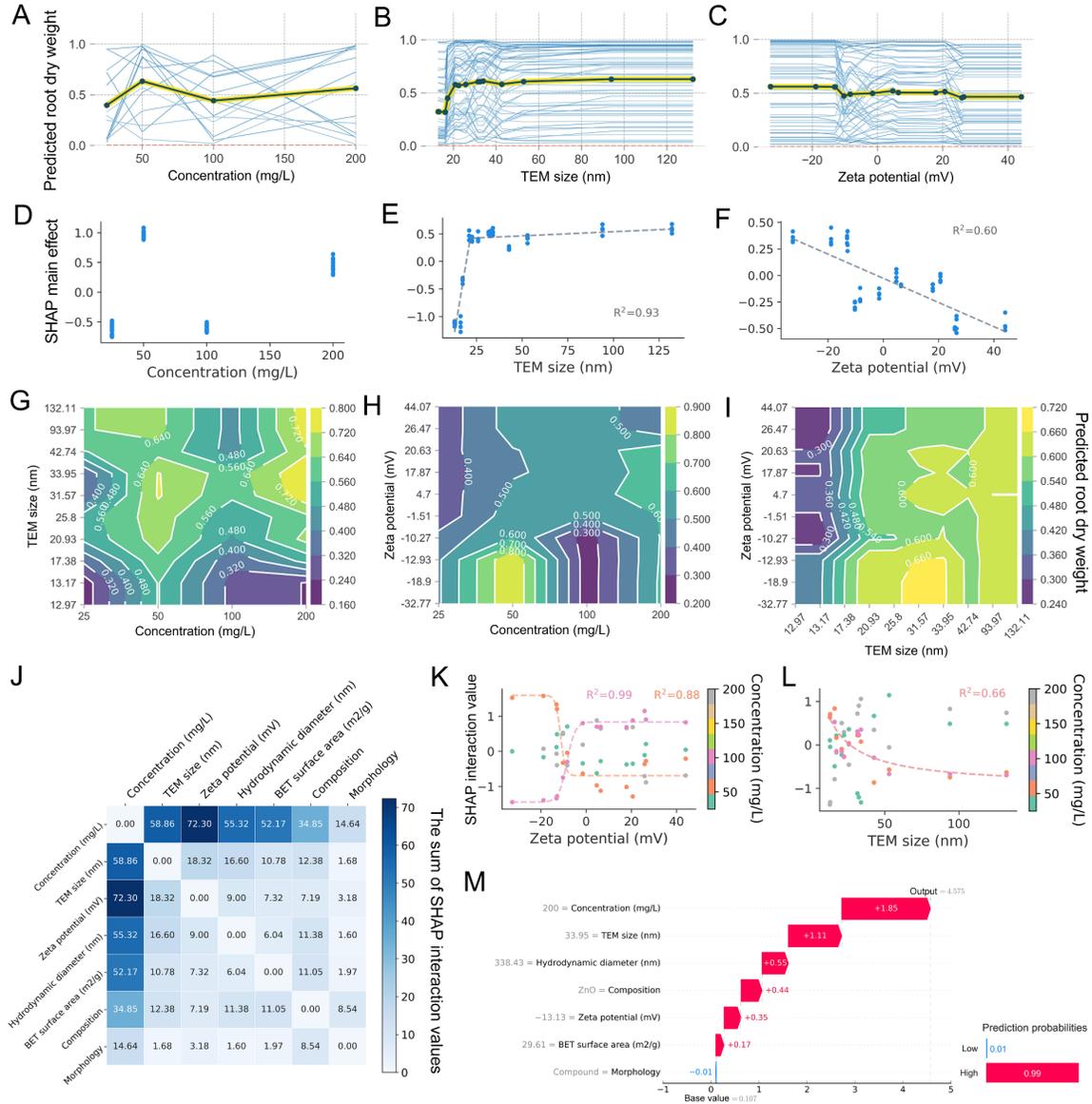

**Fig. 4.** Post hoc interpretation of the LightGBM model. A-C, PDP and ICE plots of three important features. D-F, SHAP main effects of three important features. The dashed line is the result of the piecewise linear fitting (E) and linear Fitting (F). G-I, Two-dimensional PDP between two of three important features. J, SHAP interaction values of all features. K,L, The plot of the most obvious SHAP interaction effects between two important features. The dashed line is the result of the logistic fitting (K) and plateau fitting (L) for points with the concentration of 50 mg/L and 200 mg/L respectively (K) and together (L). m, A local explanation by assigning a numeric measure of credit to each input feature. The fitting equations are provided in Supplementary method S4.

**3.5 Model-based and post hoc interpretation of the self-interpretable model**

The important features identified by post hoc interpretation of LightGBM models were used for decision tree and RuleFit modeling to simplify the problem. The AUROC scores on the test set for the decision tree and RuleFit models based on the fifth dataset split were 0.82 and 0.81 respectively (Fig. S18), which were no less than that in the LightGBM model (0.80). The decision tree is one of the easiest and most popular classification algorithms, which provides contrastive and easy-to-understand explanations by comparing an instance with decision nodes. The first leaf node was derived from the first decision node: if the zeta potential was less than -25.835 mV, 11 out of 13 instances were classified into the high level in root dry weight (Fig. S19), which was consistent with Fig. 4C,F. Besides, the second leaf node was the combined statement of TEM size and zeta potential: if the zeta potential was less than -25.835 mV and the TEM size was over 113.04 nm, eight of ten instances were classified into the high level in root dry weight, which was consistent with Fig. 4I. The RuleFit algorithm is the combination of the tree ensemble model and the sparse linear model. Therefore, it has a strong interpretation and usually performs well on complex problems. The most important rule was the conjoint statement of concentration and zeta potential (Table S6): if the concentration was more than 37.5 mg/L and less than 75 mg/L and the zeta potential was less than 5.565 mV, there was a step increase of 2.2548 for root dry weight, which was consistent with their PDP combined effects (Fig. 4H). Besides, the second important rule was consistent with the PDP combined effects between concentration and TEM size (Fig. 4G). However, it was hard to explain the overall effects of one feature in the decision tree and RuleFit models because most leaf nodes went through many decision nodes and the effects of concentration in the RuleFit model were distributed among 29 rules in 38 rules.

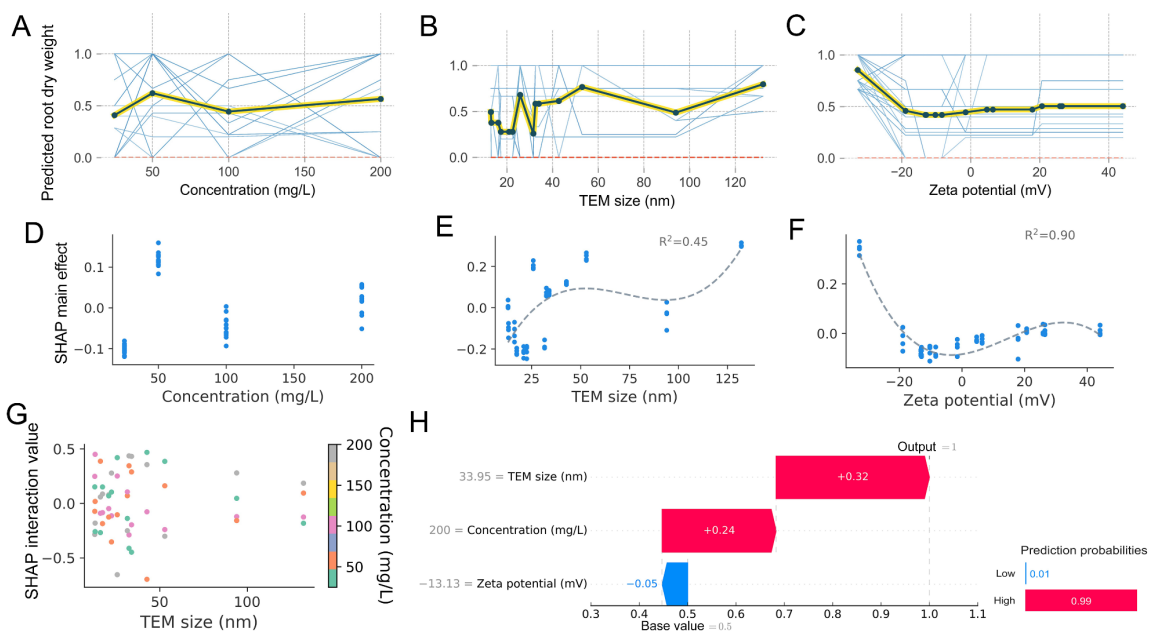

**Fig. 5.** Post hoc interpretation of the established decision tree models. A-C, PDP and ICE plots of three important features in the decision tree model. D-F, SHAP main effects of three important features in the decision tree model. The dashed line is the result of the polynomial fitting. G, The plot of the most obvious SHAP interaction effects between concentration and TEM size in the decision tree model. H, A local explanation by assigning a numeric measure of credit to each input feature in the decision tree model.

Post hoc interpretation was therefore employed to interpret the decision tree model intuitively (Fig. 5). The effect of concentration was consistent with that in the LightGBM model, but the effects of TEM size and zeta potential had some changes. This may be due to the following reasons: (i) the lack of smoothness in the decision tree algorithm; (ii) the limited feature combinations used for modeling. More specifically, small changes in input features could have a large impact on the predictions because of the existence of decision nodes; although highly correlated features have been eliminated before modeling, specific values between every two features may appear in pairs. Therefore, the feature effects in the decision tree model may be a combination of the feature effects in the LightGBM model because only three features were used in the decision tree model. Nevertheless, the general feature effects of TEM size and zeta potential (Fig. 5E,F) were similar to those in the LightGBM model (Fig. 4E,F): larger TEM size and lower zeta potential had higher root dry weight. Concentration and size had the strongest interactions but the dispersion was not obvious (Fig. S20 and Fig. 5G). The local interpretation for the ZnO nanopriming treatment with the highest SRI is shown in Fig. 5H.

**3.6. Online interactive website for prediction-level interpretation**

Prediction-level interpretation can provide detailed insights about individual predictions, which may be attractive and useful for researchers focused on specific nanomaterials. An interactive website is a great way to achieve prediction-level interpretation for all instances in the used dataset, even custom instances. We deployed the established models and the local interpretation on a website (https://seed-nanopriming-isar.streamlit.app/), which can achieve model prediction and prediction-level interpretation in real time (Fig. 6 and Fig. S21). It is worth noting that custom instances may be invalid since they ignore feature distribution and feature correlation. Moreover, local interpretable model-agnostic explanation (LIME) was employed to compare with SHAP local interpretation method on the website (Ribeiro et al., 2016).

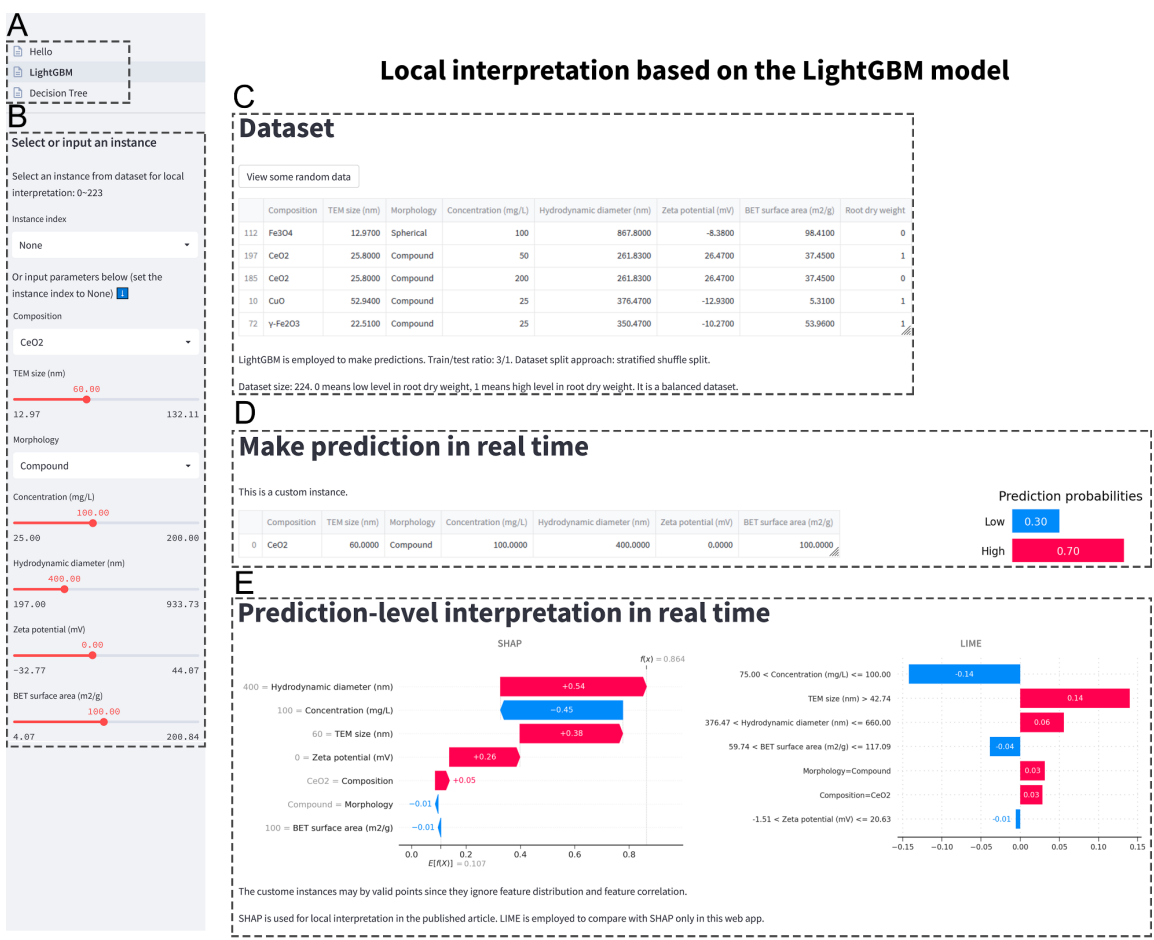

**Fig. 6.** The online interactive website for prediction-level interpretation (https://seed-nanopriming-isar.streamlit.app/). A, Navigation to different pages on this website. Hello: welcome page. LightGBM: local interpretation in the LightGBM model. Decision tree: local interpretation in the decision tree model. B, Select an instance from the used dataset or customize a sample. C, Dataset introduction and some random instances. D, Show the selected/custom instance and make a prediction. E, Prediction-level interpretation for this prediction.

## 4. Discussion

Crop plants are exposed to various environmental stresses at different growth stages. Strategies that could enhance crop resilience to different environmental stresses may generate multiple benefits in the future. Seven selected nanopriming treatments significantly increased the SRI by 13.9% and 12.6% on average under salinity and combined heat-drought stress, respectively. The treatment using ZnO (30 nm at 200 mg/L) had the highest SRI under both salinity and combined heat-drought stress, with a low nanoparticle cost of ~¥ 13.55/ha (around $ 2/ha, Supplementary method S5). This may be due to the following reasons: zinc is the only metal represented in all six enzyme classes, and acts as an important component for essential enzymes of plants and several transcription factors (Auld, 2001; Prasad et al., 2012); half of the global cereal-growing soils had Zn-deficiency problems (Liu et al., 2020), and maize was considered highly susceptible to Zn deficiency (Noulas et al., 2018); and it has been indicated that the ZnO nanoparticles could be biotransformed into the forms usable by plants (Da Cruz et al., 2017). Besides, root dry weight was the most obvious improvement among all biological endpoints, demonstrating that seed nanopriming may improve maize stress tolerance by promoting water and nutrient uptake by roots, which were also indicated in previous literature reports (El-Badri et al., 2021; Mahakham et al., 2017). It was demonstrated that Zn-containing nanoparticles could enhance root length more than other metal-containing nanoparticles in a meta-analysis (Guo et al., 2022). Although only seedling stage was investigated in this study, deep and developed roots might be essential for plant adaptation and productivity under environmental stresses in all growth stages (Giordano et al., 2021; Varshney et al., 2021). Moreover, the advances in the green synthesis of nanoparticles would enhance their biocompatibility for biotechnology applications (Shafey, 2020; Sharma et al., 2021).

Metabolomics demonstrated that maize leaves after seed nanopriming underwent many metabolic reprogrammings under both salinity and combined heat-drought stresses. Tryptophan is the biosynthetic precursor for melatonin, which could enhance plant resistance to salinity stress by directly clearing reactive oxygen species and indirectly enhancing antioxidant enzyme activity (Li et al., 2019). Secondary metabolites are recognized as natural tools that can help plants tolerate environmental stresses by regulating the gene transcriptions involved in stress tolerance (Jan et al., 2021). The high-level accumulation of benzenoids and carbohydrates and carbohydrate conjugates was observed in the SN treatment. It was reported that benzenoids serve as chemical signals and precursors of natural products related to plant stress fitness (Tinte et al., 2022). The homeostasis of carbohydrates, important sources of carbon and energy during plant metabolism, is critical for plant environmental stress tolerance. Amino sugar and nucleotide sugar metabolism was significantly enriched under both salinity and combined heat-drought stresses due to the upregulation of Glucosamine 6-phosphate and D-Glucosamine. Asparagine plays an important role in nitrogen storage and transport in plants, and it accumulates under stress conditions for the maintenance of osmotic pressure (Lea et al., 2007). However, it was down-regulated in the HdN treatment, probably due to HdN upregulating other metabolites to balance osmotic pressure and help plants survive dehydration, such as fatty acids and conjugates (Gu et al., 2020). The effect on aminoacyl-tRNA biosynthesis may indicate that HdN could mitigate combined heat-drought stress by regulating the translation process (Wang et al., 2021).

Collecting data from experiments is resource-intensive and time-consuming, underscoring the importance of making full use of existing data. In this study, we used a more rigorous performance evaluation, randomly dividing the dataset ten times (random seed from 1 to 10) and building ten models to determine the average model performance. This stringent approach may lead to a reduced model performance, but the performance assessment was unbiased. Because the model trained on a specific training set could lead to biased performance estimates, especially in a small dataset (Vabalas et al., 2019).

PMP interpretation workflow was proposed to highlight the strengths and make up for the weaknesses of different interpretive methods. Three types of ISAR were presented including the post hoc interpretation of the LightGBM model, the model-based interpretation of self-interpretable models, and the combination of model-based and post hoc interpretation of the decision tree model. Researchers and policymakers could choose which interpretation type they prefer to accept. For example, the model-based interpretation of self-interpretable models can be trusty for high-stakes decision making (Rudin, 2019). However, the post hoc interpretation of complex models with good performance may be useful for researchers to extract hypotheses from high-dimensional spaces. For example, the effects of nanoparticle size and zeta potential along with their interactions need more attention as most studies focused on concentrations.

Data-driven research with interpretable tools could accelerate the discovery and assessment of nanomaterials, while the data gap is always a challenge. Although data-sharing practices have received attention in the last decade (Yu et al., 2021), the available data in the fields of agriculture are still lacking due to the expensive data collection, large feature space, and complex data curation and quality evaluation. The complete characterization of nanomaterials, seeds/plants, environment, and experimental protocols will improve the reusability of experimental data.

**5. Conclusions**

This study presented a systematic framework to investigate the biological effects and underlying mechanisms of seed priming using low-cost nanoparticles. Seven selected nanopriming treatments significantly improved plant tolerance under both salinity and combined heat-drought stresses. Metabolomics analysis revealed that ZnO nanopriming treatment, with the highest SRI value, enhanced seedling growth under different abiotic stresses through a variety of metabolic reprogrammings. Furthermore, we introduced a promising ISAR approach that integrates various machine learning interpretation methods to gain insight into the mechanism of seed nanopriming. Concentration, size, and zeta potential of nanoparticles were identified as dominant factors, and their effects and interactions were described. To allow interested readers to understand more about the specific nanopriming treatment, we made the ISAR models and their prediction-level interpretation available on an interactive website. Additionally, the presented ISAR is a general and user-friendly technique, and could be applied to the broad fields of agricultural and environmental applications.

**Acknowledgments**

This research was supported by Zhejiang Provincial Natural Science Foundation of China under Grant No. LZ23C200005, National Natural Science Foundation of China under Grant No. 61873231, and China Scholarship Council under Grant No. 202206320340.